\documentclass[preprint,12pt, a4paper]{elsarticle}
\usepackage{epsfig}
\usepackage{amssymb}
\usepackage{hyperref}
\usepackage{url}
\usepackage{placeins}
\usepackage{caption}
\usepackage{amsthm}
\usepackage{lineno}


\journal{SoftwareX}

\begin{document}
\renewcommand{\labelenumii}{\arabic{enumi}.\arabic{enumii}}

\begin{frontmatter}

\title{SALSA: Semi-Autonomous Literature Summarization Assistant}

\author{William Schertzer}
\author{Sonakshi Gupta}
\author{Rampi Ramprasad}
\address{School of Materials Science and Engineering, Georgia Institute of Technology,
771 Ferst Drive, J. Erskine Love Building,
Atlanta, Georgia 30332-0245, United States}

\begin{abstract}
SALSA (Semi-Autonomous Literature Summarization Assistant) is an open-source, human-in-the-loop platform for extracting structured scientific datasets from multimodal literature sources. The software combines document parsing, large language models, optical character recognition, computer vision, figure digitization, and user-guided correction tools to recover structured information from text, tables, figures, and captions. Users can configure extraction stages, define dataset schemas, perform interventions on digitized figures, and export verified data for downstream analysis and machine learning. SALSA is designed to automate repetitive literature curation tasks while preserving oversight where expert judgment is required. By supporting customizable extraction workflows across diverse input types, the software provides a flexible framework for scalable, reliable data curation for materials research, and potentially across several disciplines. Users are responsible for ensuring that all inputs, extraction workflows, and downstream uses comply with applicable publisher agreements, copyright and licensing terms, institutional policies, and data-use requirements.
\end{abstract}


\end{frontmatter}

%\linenumbers

\section*{Metadata}
\label{}

\begin{center}
\label{tab:code-metadata}

\makebox[\textwidth][c]{%
\begin{tabular}{|l|p{6.5cm}|p{6.5cm}|}
\hline
\textbf{Nr.} & \textbf{Code metadata description} & \textbf{Metadata} \\
\hline
C1 & Current code version & v1.0 \\
\hline
C2 & Permanent GitHub link to code/repository used for this code version & \url{https://github.com/Ramprasad-Group/SALSA}\\
\hline
C3 & Legal Code License & Apache License 2.0 \\
\hline
C4 & Code versioning system used & git \\
\hline
C5 & Software code languages, tools, and services used & Python 3.8, Conda, Streamlit \\
\hline
C6 & Compilation requirements, operating environments \& dependencies & Local deployment requires a Linux operating system, dependencies listed under environment.yml and local setup instructions listed in the README.md \\
\hline
C7 & Link to developer documentation/manual & Video 1, Supplementary Information \\
\hline
C8 & Support email for questions & wschertzer3@gatech.edu \\
\hline
\end{tabular}%
}
\end{center}
\captionof{table}{Code metadata}

\clearpage

\section{Motivation and significance} 

The volume of digitally accessible data has expanded dramatically in recent decades as a result of advances in digitization, FAIR data practices, high-throughput experimentation, simulation, and open-access publishing \cite{wilkinson_fair_2016, bornmann_growth_2015, sagiroglu_big_2013}. In parallel, the shift towards data-driven scientific discovery has increased the demand for datasets that are sufficiently structured, contextualized, and traceable for statistical analysis and machine learning. The resulting bottleneck is no longer simply access to information; it is the conversion of information written for human communication into records that can be used reliably by computational workflows.\\

This problem is especially prevalent in chemistry and materials research. A reported property value is rarely meaningful in isolation; its interpretation may depend on the associated material identity, composition, processing history, measurement method, and testing environment \cite{meliante_evaluation_2025, jansen_data_2025, jensen_chatgpt-4o_2025}. These elements are often distributed across narrative text, tables, figures, captions, and supplementary files rather than presented as a unified record \cite{jansen_data_2025, jensen_chatgpt-4o_2025, gupta_data_2024, he_zeoreader_2025}. Although these heterogeneous sources contain information that could support statistical analysis and machine learning, their value can only be realized after the relevant data and their distributed scientific context have been identified, connected, and converted into structured, traceable records.\\

Manual data curation remains one of the most common ways to address this bottleneck \cite{he_zeoreader_2025, torre-lopez_artificial_2023, stocker_rethinking_2025, chen_chemminer_2024}. When performed carefully, manual extraction can produce high-quality datasets that preserve context; however, it is labor-intensive, difficult to scale, and frequently requires highly trained domain experts to spend substantial time on repetitive data collection rather than interpretation, modeling, experiment design, or theory development \cite{ stocker_rethinking_2025, he_towards_2026, tran_design_2024}. Fully automated extraction tools can improve throughput, but they can also produce errors through omission or errors that are difficult to detect because the extracted values may appear structurally valid while being factually incorrect \cite{jansen_data_2025, jensen_chatgpt-4o_2025, gupta_data_2024, peng_accuracy_2025,polak_extracting_2024}.\\

Existing tools address individual components of the document-to-data pipeline, including text analysis, chemical synthesis extraction, chart digitization, systematic-review assistance, and autonomous dataset construction \cite{gupta_data_2024, jansen_data_2025, jensen_chatgpt-4o_2025, he_zeoreader_2025, chen_chemminer_2024, jiang_plot2spectra_2022, wang_scidasynth_2025, roy_beyond_2026, radanliev_review_2023, lee_autonomous_2024}. However, many of these tools are designed for a single data modality, such as text, tables, figures or for a narrowly defined application (e.g. extracting zeolite synthesis conditions \cite{he_zeoreader_2025} or extracting values from line charts depicting spectral data \cite{jiang_plot2spectra_2022}). Their extraction models, instructions, and output schemas may also be fixed or difficult to modify, limiting their transferability to new document collections and research questions. In addition, many operate as scripts or isolated services without a graphical interface through which domain experts can configure the workflow, inspect intermediate outputs, correct extraction errors, and assemble the resulting information into a unified dataset. Consequently, researchers often must combine multiple specialized programs and manually transfer information between document acquisition, relevance screening, content extraction, contextualization, validation, and final dataset construction.\\

SALSA, the Semi-Autonomous Literature Summarization Assistant, was developed to connect these operations in an inspectable and iteratively improvable human-AI workflow. Its main contribution is not a single extraction model, but an integrated interface in which users can configure document inputs, select processing stages, choose hosted or local model backends, edit LLM prompts, define dataset schemas, inspect intermediate evidence, intervene in figure digitization, and rerun individual stages. The platform is not intended to replace expert judgment. Instead, it automates repetitive extraction tasks, organizes intermediate evidence, and provides interactive tools for expert verification and correction.\\

Although its initial use case focuses on materials science literature, the architecture is not restricted to published articles or a single scientific domain. The same configurable workflow can be applied to other scientific and technical documents when appropriate extraction instructions and dataset schemas are provided. This broader scope is important because published literature represents only a fraction of the data available within industrial, institutional, and laboratory archives. Historical experiments, negative or inconclusive results, process notes, characterization summaries, and other technical records may instead be distributed across internal reports, presentations, spreadsheets, scanned documents, laboratory notebooks, and isolated images. By supporting heterogeneous document inputs and user-defined extraction objectives, SALSA can help convert these otherwise fragmented records into structured and traceable datasets while preserving the expert oversight needed to interpret domain-specific context.\\

The software is distributed for local deployment so that users can control document acquisition, intermediate outputs, model credentials and backends, and data storage/retention policies while adapting the software to their own access permissions, security requirements, and computational resources. Use of the software does not grant permission to access, copy, redistribute, mine, or reuse documents beyond the rights they already possess. Users are responsible for ensuring that their inputs, extraction workflows, model calls, stored outputs, and downstream uses comply with applicable publisher agreements, copyright and licensing terms, institutional policies, data-use agreements, confidentiality requirements, and privacy regulations.

\section{Software description} 

SALSA is a flexible, expert-guided, AI-assisted platform designed to support a thorough, robust and continuously improvable  literature data curation workflow, beginning with scientific document input and ending with human-verified structured datasets.

\subsection{Software architecture}

\begin{figure*}[t]
    \centering
    \includegraphics[width=\textwidth]{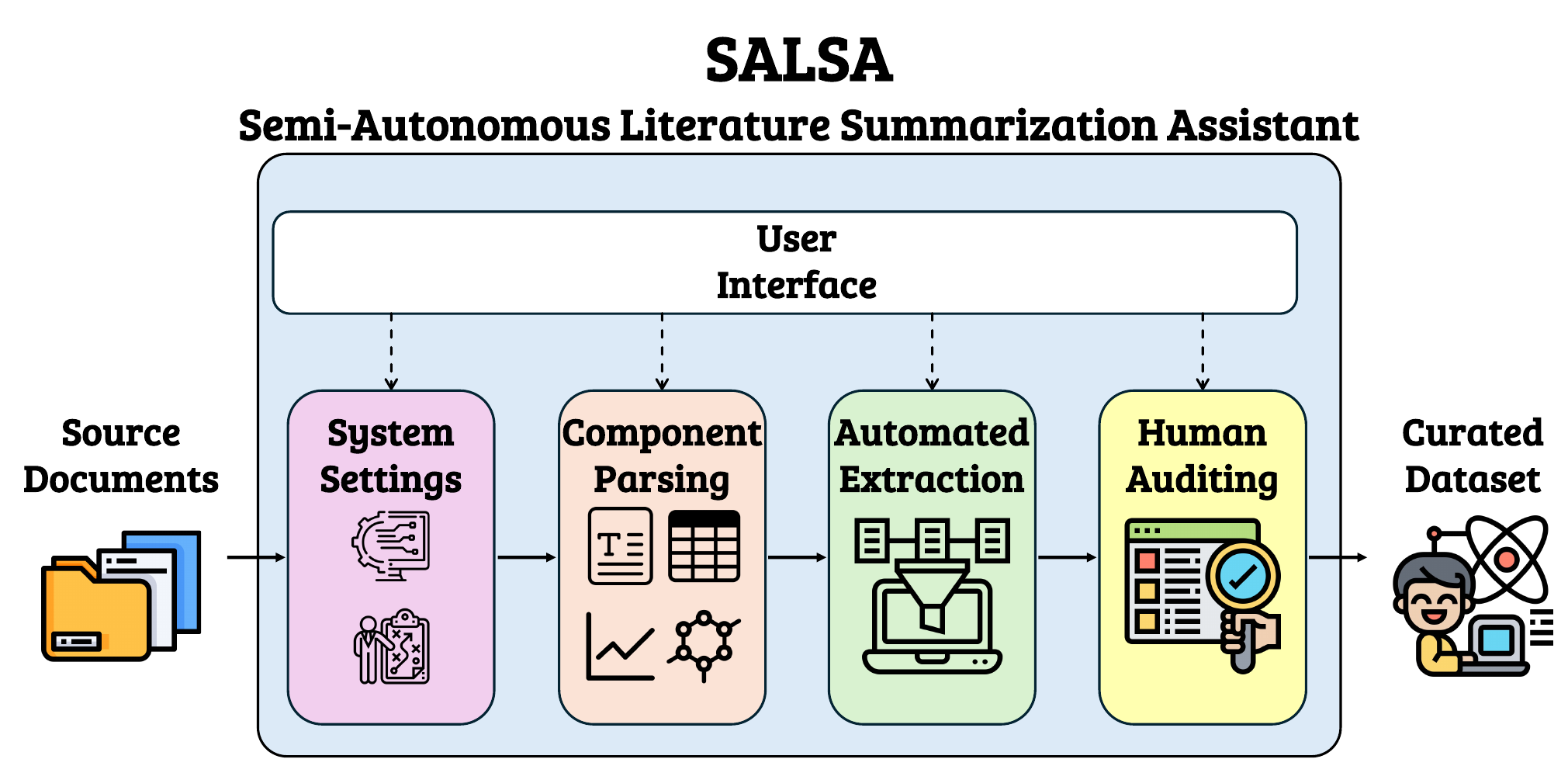}
    \caption{Architecture of the SALSA software}
    \label{fig:architecture}
\end{figure*}

Figure \ref{fig:architecture} summarizes the software architecture. Inputs are entered through a browser-based interface either as a list of DOIs, or as a collection of PDFs or standalone figure image files.  The interface manages workspaces, run configurations, stage selection, model settings, extraction instructions, target schemas, and user interventions. An orchestration layer validates the run configuration, routes each input into the enabled modules, and writes stage outputs to a persistent run directory. Task-specific modules provide document parsing, table and figure extraction, relevance filtering, figure digitization, chemistry and sample-context extraction, and dataset construction. Upon completion, each run retains the configuration, parsed objects, model responses, digitized values, user corrections, and exported datasets for review.\\

Model-dependent operations use a shared provider interface. The current implementation supports OpenAI endpoints and local Ollama servers. Additional model backends can be easily integrated via the source code.

\subsection{Implementation and repository organization}

\begin{figure*}[t!]
    \centering
    \includegraphics[width=\textwidth]{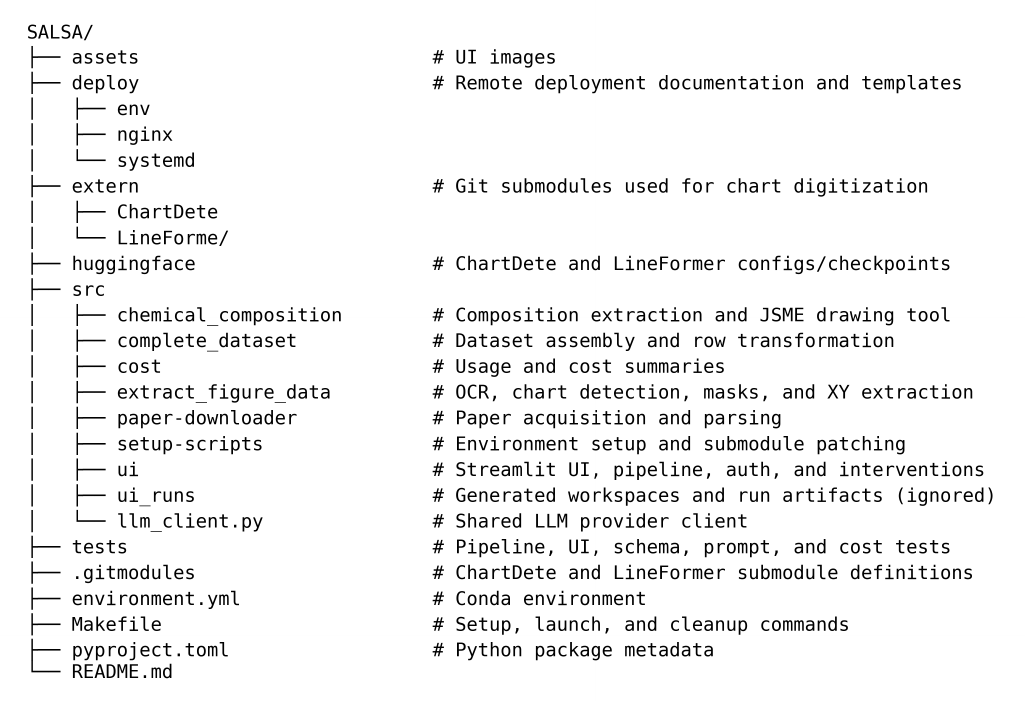}
    \caption{File tree of the SALSA repository, indicating the key components of the code including ui, modular functionalities, tests, setup and requirements. }
    \label{fig:repo_tree}
\end{figure*}

The Python source code is organized as shown in Figure \ref{fig:repo_tree}. Core functionalities are implemented as separate modules within \texttt{src/}. This directory also contains \texttt{llm\_client.py}, which provides a shared abstraction layer for hosted and local model backends. The remaining files support installation, configuration, and deployment, as well as the integration of the external ChartDete and LineFormer submodules \cite{10.1007/978-3-031-41676-7_13, lal_lineformer_2023} and associated Hugging Face model weights.\\

\subsection{Software functionalities}

Custom settings can be configured either for broad document inspection or for targeted data extraction into a domain-specific schema. General operations (e.g. document parsing, object filtering, figure digitization, and tabular assembly) are applicable across scientific fields. Context-aware chemical structure recovery is an additional capability intended for chemistry and materials applications. Figures S1-S4 provide an overview of the available features, and video 1 and figures S5-S9 shows these features being used practice.

\subsubsection{Document parsing and multimodal extraction}

DOI-based runs retrieve article content through the configured publisher API keys. The same downstream workflow can be applied to user-uploaded PDFs when automated retrieval is unavailable or not permitted. Standalone image files are accepted for processing figures.\\

Document parsers separate an article into text, tables, figures, captions, and source metadata. Tables are reconstructed into row-column representations, and figures are stored as image objects linked to associated captions. These objects are passed selectively to subsequent stages.

\subsubsection{Table and figure filtering}

A relevance stage reduces the set of tables and figures sent to more expensive extraction modules. Selection is based on user instructions together with captions, reconstructed content, and visual content when a vision-capable model is enabled (for figures). The criteria can describe a property, experimental method, sample class, or any other evidence specified by the user.

\subsubsection{Figure digitization and user intervention}

The software provides two complementary routes for extracting data from plots. The optical character recognition (OCR) route uses the open-source ChartDete Python package \cite{10.1007/978-3-031-41676-7_13} to identify plot regions, axes, and tick labels and to calibrate pixel coordinates against numerical axis values. The resulting calibration is passed to the open-source LineFormer Python package \cite{lal_lineformer_2023}, which detects plotted series and associates them with their corresponding legend entries, marker styles, and colors. The detected pixel coordinates are then converted into numerical values. Alternatively, figures can be passed to a vision-language model to interpret figures directly. \\

User-intervention tools are available for figures processed through the OCR route. Scientific figures frequently contain overlapping markers, low-resolution text, multi-panel layouts, nonstandard axes, or ambiguous legends that can introduce errors at different stages of the computer-vision pipeline. The software's graphical interface allows users to crop figures, correct axis titles and calibrate limits, edit series labels, and override or create series masks. As vision-language models do not provide intermediate outputs for chart elements, their results cannot be modified via user interventions.

\subsubsection{Chemical composition and context extraction}

In the fields of chemistry and materials science, many datasets require linking property values to material identities, sample labels, chemical compositions, formulation details, processing conditions, and testing environments. The software supports this task through a two-stage chemical composition and sample-context extraction workflow. In the first stage, relevant contextual evidence is collected from text, tables, captions, figures, and metadata. In the second stage, this evidence is used to extract structured composition records according to the user-defined schema.\\

Depending on the extraction objective, these records may include sample names, polymer or molecular components, component ratios, additives, solvents, concentrations, synthesis conditions, treatment conditions, testing environments, and other domain-specific descriptors.

\subsubsection{Schema-driven dataset generation}

The dataset-generation stage consolidates information from the enabled extraction modules into a user-defined tabular schema. Users specify the desired dataset columns, default values, formatting rules, and extraction instructions. Then, evidence from parsed text, reconstructed tables, extracted figures, digitized coordinates, captions, chemical composition records, and source metadata are combined to generate structured dataset rows.\\

\subsection{Practical usage considerations}

Model-backed stages can be assigned to a hosted OpenAI-compatible endpoint or to an Ollama server. A local model may be preferred when documents cannot be transmitted outside an institutional environment, when repeated extraction would otherwise incur substantial API cost, or when users require direct control over model versions. Hosted models may provide stronger multimodal or long-context performance, but they introduce provider-specific costs, retention policies, and data-use terms. Local inference likewise requires adequate memory and compute, and its extraction quality depends on the selected model.\\

Users should begin with a small, representative document set, inspect intermediate outputs, and refine prompts and schemas before scaling. Extraction quality is affected by document quality and layout, figure resolution, model capability, context length, and schema complexity. No model choice eliminates the need for expert validation, particularly when the resulting dataset will support scientific claims or produce machine-learning models.

\section{Illustrative examples}

The following examples illustrate representative use cases of SALSA for scientific literature data curation. The examples are intended to demonstrate the major capabilities of the software rather than serve as exhaustive domain-specific case studies. A walkthrough of these workflows is provided in Video 1 and figures S5-S9.

\subsection{General scientific document processing}

Ten highly cited articles identified in a 2025 Nature survey were selected to exercise different publishers, layouts, and scientific content types \cite{van_noorden_these_2025}. The general workflow was configured to preserve article text, reconstructed tables, extracted figures, captions, and metadata without applying any application-specific extraction instructions or dataset schemas.\\

This example demonstrates a practical aspect of semi-autonomous data extraction: access routes vary substantially across journals and document types. DOI-based processing was used when the article could be retrieved through an available, authorized API. When publisher policies or API limitations prevented automated retrieval, the identical parsing workflow was applied to a PDF supplied by the user. In both cases, the output was an organized corpus of text, tables and figures ready for further processing and analysis.

\subsection{Anion exchange membrane degradation}

The second example demonstrates SALSA's ability to curate a highly contextual, domain-specific dataset from dispersed information sources within scientific publications. The objective is to extract anion exchange membrane degradation data, with a focus on hydroxide conductivity as a function of aging time and testing conditions.\\

The target dataset schema contains necessary fields for training machine learning models as described in our previous work \cite{schertzer_ai-assisted_2026}, namely: sample name, material identity, membrane composition, aging time, conductivity value, temperature, relative humidity, alkaline solution identity and concentration, stability testing temperature, and source publication information. The table and figure filtering stage is then configured to retain elements relevant to conductivity, degradation, alkaline stability, aging conditions, and membrane composition.\\

Degradation curves are digitized, after which the user checks the plotting region, axis calibration, series labels, and extracted coordinates. The material and experimental context of each series are identified by passing the text and table evidence to the specified LLM. For example, a legend label can be linked to a formulation described in the experimental section and to an aging condition reported in a caption or table.

\section{Impact}

Expert judgment remains essential to the creation of high-quality scientific datasets, but purely manual data curation cannot scale with the growing volume and diversity of available information. In many fields, and especially in industrial materials research, valuable data is distributed across an ever-increasing corpus of articles, figures, supplementary files, scanned records, laboratory notebooks, internal reports, spreadsheets, and other document types. Constructing an application-specific dataset therefore often requires researchers to locate, interpret, organize, extract, and validate information using a fragmented collection of tools and manual procedures. These processes are time-consuming and frequently require expertise in both the scientific domain and the exact structure of the source documents. SALSA addresses this bottleneck by extending the scale at which expert judgment can be applied without removing that judgment from the curation process.\\

The software automates repetitive extraction and processing tasks while allowing users to define dataset objectives and schemas, configure models and extraction instructions, inspect source evidence, correct intermediate outputs, and rerun individual processing stages. This shifts the role of the expert from manually performing every operation to guiding, evaluating, and intervening in an automated workflow. Further, by integrating these operations within a common framework, the organizational burden associated with coordinating disparate extraction tools, file formats, and manually maintained records is significantly reduced.\\

The framework preserves not only the resulting dataset but also the procedure used to construct it. Dataset schemas, model and API configurations, extraction instructions, source evidence, intermediate outputs, and user corrections can be retained as part of the curation workflow. This allows extracted values to be traced to their underlying evidence and enables individual processing stages to be inspected, modified, or repeated as research objectives and data requirements change. Consequently, datasets can be audited and updated without requiring the entire curation process to be reconstructed.\\

\section{Caveats and Disclaimers}

SALSA is intended to support responsible, user-supervised extraction of structured data from documents that users are authorized to access and process. The software does not provide, expand, or imply permission to access, download, copy, mine, redistribute, or reuse publisher-hosted content, copyrighted materials, proprietary records, or sensitive documents. Users remain solely responsible for ensuring that all inputs, extraction workflows, model calls, stored outputs, and downstream uses comply with applicable publisher agreements, text-and-data-mining policies, copyright and licensing terms, institutional subscriptions, data-use agreements, confidentiality requirements, and privacy regulations. Access to a document through an institutional subscription, library portal, collaborator, or internal repository should not be assumed to grant permission for automated extraction, large-scale mining, redistribution, or reuse outside the terms governing that access.\\

When processing publisher-hosted literature, users should follow the relevant publisher and institutional requirements, including the use of approved APIs, text-and-data-mining routes, or permission mechanisms where applicable. When processing internal, proprietary, confidential, personal, or otherwise sensitive documents, users should ensure that document handling, model-provider interactions, intermediate storage, and exported datasets are consistent with the governing organizational and legal requirements.\\

\section{Conclusions}

This work introduced SALSA, the Semi-Autonomous Literature Summarization Assistant, an open-source, human-in-the-loop platform for constructing structured scientific datasets from heterogeneous documents. The software integrates document ingestion, multimodal extraction across text, tables, figures, and chemical composition records, configurable models and extraction instructions, user-defined schemas, and interactive review tools within a single workflow. These capabilities allow evidence distributed across different sources and data modalities to be connected while retaining the context required for scientific interpretation.\\

SALSA is designed to make expert judgment more scalable rather than replace it. The software automates repetitive processing stages while allowing users to inspect source evidence and intermediate outputs, correct errors, refine extraction instructions, and rerun selected operations. Preserving model configurations, instructions, intermediate outputs, user corrections, and final datasets also provides the information needed to audit, reproduce, and update data-curation workflows. This combination of automation, expert oversight, and traceability is particularly important for scientific machine learning, where model reliability depends directly on the quality and consistency of the underlying data.

\section*{Acknowledgements}
This work was supported as part of the UNCAGE-ME, an Energy Frontier Research Center funded by the U.S. Department of Energy, Office of Science, Basic Energy Sciences at the Georgia Institute of Technology under award \#DE-SC0012577.

\clearpage

\label{}

\bibliographystyle{elsarticle-num} 
\bibliography{bibliography}

\end{document}